\documentclass[preprint,12pt]{elsarticle}
\usepackage{amssymb}
\usepackage{graphicx}
\usepackage{float}
\usepackage{subfigure}
\usepackage{amsmath}
\usepackage[pagewise]{lineno}
\usepackage{algorithm}
\usepackage{algorithmic}
\usepackage{booktabs}
\usepackage{multirow}
\usepackage{bm}
\usepackage[cmintegrals]{newtxmath}
\usepackage[colorlinks,
linkcolor=blue,
anchorcolor=green,
citecolor=red,
]{hyperref}

\journal{Transportation Research Part C}

\makeatletter
\newenvironment{breakablealgorithm}
{
	\begin{center}
		\refstepcounter{algorithm}
		\hrule height.8pt depth0pt \kern2pt%
		\renewcommand{\caption}[2][\relax]{
			{\raggedright\textbf{\ALG@name~\thealgorithm} ##2\par}%
			\ifx\relax##1\relax 
			\addcontentsline{loa}{algorithm}{\protect\numberline{\thealgorithm}##2}%
			\else 
			\addcontentsline{loa}{algorithm}{\protect\numberline{\thealgorithm}##1}%
			\fi
			\kern2pt\hrule\kern2pt
		}
	}{
		\kern2pt\hrule\relax
	\end{center}
}
\makeatother

\begin{document}
\begin{frontmatter}
\title{Network-wide Traffic Signal Control Optimization Using a Multi-agent Deep Reinforcement Learning}

\author[um]{Zhenning Li}
\author[seu]{Hao Yu\corref{cor1}}
\ead{haoyu@hawaii.edu}
\author[uhm]{Guohui Zhang}
\author[ud]{Shangjia Dong}
\author[um]{Chengzhong Xu}
\address[um]{State Key Laboratory of Internet of Things for Smart City, University of Macau}
\address[seu] {School of Transportation, Southeast University}
\address[uhm]{Department of Civil and Environmental Engineering, University of Hawaii at Manoa}
\address[ud]{Department of Civil and Environmental Engineering, University of Delaware}
\cortext[cor1]{Corresponding author}

\begin{abstract}
Inefficient traffic control may cause numerous problems such as traffic congestion and energy waste. This paper proposes a novel multi-agent reinforcement learning method, named KS-DDPG (\textbf{K}nowledge \textbf{S}haring  \textbf{D}eep \textbf{D}eterministic \textbf{P}olicy \textbf{G}radient ) to achieve optimal control by enhancing the cooperation between traffic signals. By introducing the knowledge-sharing enabled communication protocol, each agent can access to the collective representation of the traffic environment collected by all agents. The proposed method is evaluated through two experiments respectively using synthetic and real-world datasets. The comparison with state-of-the-art reinforcement learning-based and conventional transportation methods demonstrate the proposed KS-DDPG has significant efficiency in controlling large-scale transportation networks and coping with fluctuations in traffic flow. In addition, the introduced communication mechanism has also been proven to speed up the convergence of the model without significantly increasing the computational burden.

\end{abstract}

\begin{highlights}
\item A multi-agent reinforcement learning for adaptive traffic signal control optimization.
\item Consider control unit at intersection as agent that can communicate with others through knowledge-sharing protocol.
\item Proposed algorithm achieves consistent improvements over baselines on both simulated and real-world data.
\end{highlights}


\end{frontmatter}

\section{Introduction}
Traffic signal control is an efficient method of protecting traffic participants at intersections where multiple streams of traffic interact. Because of the capability of responding to fluctuating traffic demands, the adaptive traffic signal control (ATSC) system has been broadly implemented, and has attracted considerable interest in the research community \cite{VandeWeg2018}. By receiving and processing data from strategically placed sensors, the ATSC system works based on real-time traffic dynamics about travel demand, traffic conditions and performance measurements of the control logic, such as traffic volume, queue length, vehicle speed, and travel time \cite{Wang2019}. Compared to the fixed-time control, ATSC has been proved to improve the quality of service that travelers experience on roadways, especially in busy urban areas, while reducing travel time by more than 10\% on average \cite{USDOT2017}. However, a malfunctioning ATSC system may bring about more serious congestion and even traffic accidents \cite{li2019}.

Numerous interdisciplinary approaches have been applied to improve the efficiency of ATSC in the past two decades, including but not limited to fuzzy logic \cite{cheng2016}, case-based reasoning \cite{El2011}, artificial neural networks \cite{Ghanim2015}, genetic algorithm \cite{odeh2015}, and immune network algorithm \cite{Darmoul2017}. Among these approaches, reinforcement learning (RL), which takes sequential actions rooted in Markov Decision Process (MDP) with a rewarding or penalizing criterion, is a promising approach to optimize the ATSC system. It is a type of goal-oriented algorithm that learns how to achieve complex objectives in specific dimensions by learning from the experience. Different from conventional model-driven approaches that only work well under certain assumptions \cite{Hegyi2005,Liu2009} (e.g., deterministic arrival rate, unlimited road capacity), RL is able to learn how to achieve complex objectives in specific dimensions by learning from the experience of interacting with the environment. Therefore, it has the potential to solving the ATSC optimization problem without strong assumptions, and to learn good strategies directly from trials and errors \cite{Wei2019} .

\begin{figure}[tbh!]
	\centering
	\includegraphics[width=1\linewidth]{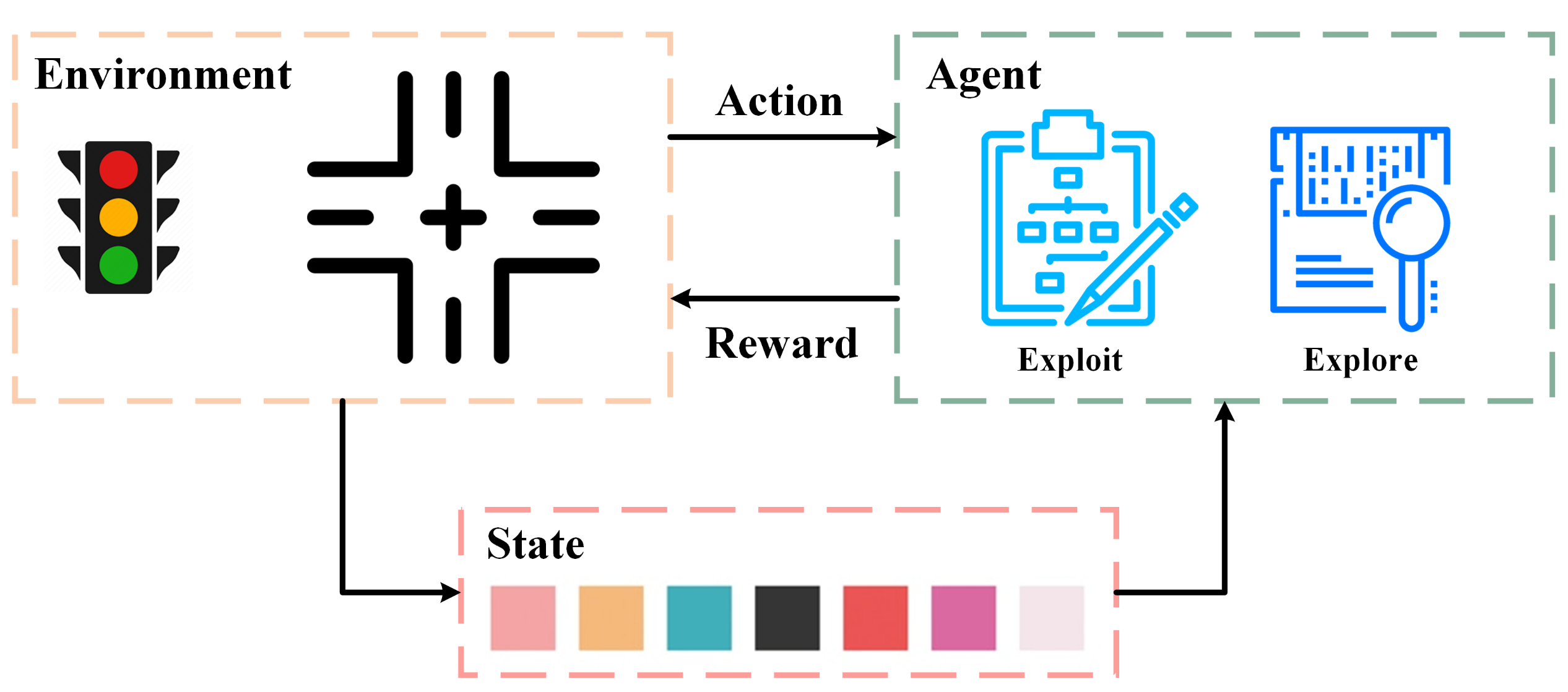}
	\caption{Reinforcement Learning Framework for Traffic Light Control}
	\label{Fig1}
\end{figure}

Let us take one of the most common settings for ATSC optimization using a single-agent RL framework in previous studies as an illustration \cite{prashanth2010}. As shown in Fig. \ref{Fig1}, the control unit in an isolated intersection is treated as an \textit{agent}, and interacts with the traffic \textit{environment} in a closed-loop MDP. The control policy is obtained by mapping from the traffic \textit{states} (e.g., waiting time, queue length, total delay, etc.) to the corresponding optimal control \textit{actions} (e.g., phase shift, cycle length change, green time extension, etc.). The agent iteratively receives a feedback \textit{reward} for actions taken and adjusts the policy until it converges to the optimal control results. During the decision process, the policy that the agent takes combines both \textit{exploitation} of already learned policies and \textit{exploration} of new policies that never met before. Studies using similar RL frameworks to manipulate ATSC are not rare in the last two decades and have provided beneficial references for research \cite{abdu2003,Arel2010,Gend2017,Mousavi2017,Zhang2020}. For instance, a single-agent model-free Q-learning algorithm was developed for optimizing signal timing in a single intersection \cite{abdu2003}.In this study, the authors used queue length as the state representation and accumulative delay between two action cycles as the reward. The comparison between the algorithm and the fixed-time control strategy under different traffic flows demonstrated the advantages of the RL-based method in dealing with fluctuating traffic flow demand. Single-agent SARSA-based algorithms using similar state-action-reward settings were also widely implemented in previous studies to solve the ATSC optimization problem of isolated intersections \cite{Thrope1996,Wen2007,Tantawy2014}.

However, many additional challenges arise when extending such a single-agent RL to a multi-agent system such as ATSC in a transportation network with multiple \textbf{interacting} intersections. First, in the commonly used centralized learning scheme, different agents are represented by a single agent and collaboratively executed towards a joint policy. The foremost drawback of this structure is that as the number of agents increases, the input size increases linearly, while the output joint policy space grows exponentially, making it difficult to tackle the network-wide ATSC optimization problem, especially in a large network. For instance, in the work of \cite{Prashanth2010}, they trained one centralized agent to determine the joint actions for all intersections in the network. However, this approach did not perform as well as expected due to the dimension curse of action space. In addition, in this kind of centralized optimization framework, collecting and processing information must be repeated to some extent, therefore it may face stability issues during deployment. Further complications of the \textit{exploration-exploitation dilemma} also arise due to the presence of multiple intersections. The exploration-exploitation trade-off requires online RL algorithms that balance between the exploitation of the agent’s current knowledge with exploratory information gathering actions taken to improve that knowledge \cite{Busoniu2010}. Agents have to not only explore information about the environment, but also the information from other control units in order to adapt to their behaviors. However, too much exploration can undermine the stability of other agents and affect the efficiency of the algorithm, making it more difficult to explore the agent’s learning tasks.  While in a decentralized structure, since multiple control units need to interact with each other during the optimization, therefore, the property in an MDP that the reward distribution and dynamics need to be stationary is violated as the reward received by an agent also depends on other agents' actions \cite{Shou2020}. This issue, known as the \textit{moving-target} problem, i.e., the best policy changes as the other agents’ policies change, eliminates convergence guarantees and introduces extra learning instabilities. It becomes more troublesome, as our environment is characterized by a partially observable feature, that is, the agent cannot fully access the entire state space, and coordination is essential \cite{Zhang2019}. In an ATSC system, the need for coordination comes from the fact that the environmental impact of any agent’s actions also depends on actions taken by other agents. Therefore, each agent's action must be consistent with others in order to achieve their expected results, e.g., minimizing network-wide delay, maximizing total throughput, etc.

In order to mitigate these issues, many previous studies have adopted various independently modeling RL frameworks, in which they treated each intersection as one agent and trained them separately \cite{Abdoos2011,Aslani2017,Xiong2019}. For instance, \cite{Mannion2016} developed three RL-based ATSC optimization algorithms and compared the control performance of them with a fixed-time control via experiments designed in a simple 3$\times$3 grid network with uniform traffic flow. In their settings, each RL agent is responsible for controlling the light sequences of a single junction and can only observe the traffic state on the lanes directly connecting with the intersection. It should be noted that many early attempts like this work did not fully take advantage of communication protocols between different agents, and they assumed the agent can only adapt to its private observations accordingly without considering any other parts of the entire environment \cite{Khamis2012}. In some simple scenarios (such as small synthetic network and arterial network), this processing method has been proven to have acceptable performance \cite{Brys2014,Prashanth2011}. However, when the network becomes more complex, the non-stationary problem can no longer be neglected. Previous studies have found that if there is no communication or coordination mechanism among agents, the learning process usually lasts significantly longer than expected, or even is unable to converge to stationary policies \cite{Nowe2012}. The underlying reason for this situation is that the agent is unable to timely respond and adapt to the environment change before the environment has already changed again.

A recently proposed algorithm, named Multi-Agent Deep Deterministic Policy Gradient (MADDPG) \cite{Lowe2017}, is a trending state-of-the-art \textit{multi-agent RL} paradigm and could partially address the above-mentioned problems encountered by single-agent RLs. This algorithm is enlightened of the actor-critic algorithm \cite{Konda2000} in which each agent has an actor to select actions and a critic to evaluate them. MADDPG utilizes a \textit{centralized learning and decentralized execution} paradigm where a group of agents can be trained simultaneously by applying a centralized method. In addition, this approach can address the emergence of environment non-stationary and has been shown to perform well in a variety of mixed competitive and cooperative environments \cite{Wang2020,Qie2019}. However, MADDPG is still not the ideal panacea for our problem.  First, as the studied network-wide ATSC optimization is a partially observable and sequential multi-agent decision-making problem,  each agent is unable to observe the underlying Markov state. Taking into account the fact that traffic traverses the network from one intersection to another, cooperation among intersections is essential to help the agent have a better understanding of the entire environment. Based on the engineering experience, it is quite common that a well-functioning phase setting at this intersection without considering the state of the entire traffic environment may lead to significant congestion at other intersections after a period of time. In addition, if each agent only takes its optimal actions based on its own observation, in order to adapt to fluctuating traffic flows, the agent may need to constantly shift its phase setting, which may, unfortunately, result in the control system being less robust and effective.  Numerous studies have also proven that considering cooperation during ATSC optimization could reduce phase adjustment times and achieve better control performance (e.g., larger average speed, less delay, etc.)  \cite{Wei2019, Arel2010, Chu2019}. Since intersections are able to interact, developing an efficient communication protocol to allow agents to share their observations with others may be a promising way to achieve cooperation and collaboration among traffic signals. However, under the MADDPG framework, agents can only share each other's actions and observations during training through the critics, and there are no means to develop an explicit form of communication through their experiences.

In order to address all the aforementioned issues, based on the MADDPG framework, we further propose a novitiate multi-agent RL algorithm allowing agents to efficiently communicate, named KS-DDPG (\textbf{K}nowledge \textbf{S}haring-\textbf{D}eep \textbf{D}eterministic \textbf{P}olicy \textbf{G}radient). In our setting, communication among agents is realized via a carefully-designed  \textit{knowledge sharing mechanism}. The knowledge that represents the collective understanding of the environment by all agents is stored in the shared \textit{knowledge container}. It is a highly compact vector with a pre-defined capacity and is constantly updated condition of the local observation history of each agent during the modeling process through non-linear gating-based operations. With the knowledge sharing mechanism, each agent could capture relevant aspects of the environment, and interpret and reconstruct the shared knowledge in its own way. The agent makes its decision based on not only its private observation but also its unique understanding of the shared knowledge obtained via the communication protocol. As far as we are aware, this is the first paper using such techniques. The remainder of this paper is organized as follows. Section 2 reviews related work on MDP and reinforcement learning. Section 3 presents the formalization of the problem setup, the details of the proposed method, and the description of the learning process. In Section 4, the proposed method is evaluated by a grid simulated network and a real-world traffic network in Montgomery County, Maryland, respectively. Section 5 summarizes this paper and discusses future research directions.

\section{Related Work}
In this section, we will introduce concepts and terminology that are related to our work, including MDP, classical reinforcement learning algorithms, and the MADDPG algorithm, respectively.

\subsection{Multi-agent Extension of Markov Decision Process}\label{MDP}
The network-wide ATSC optimization problem can be described as a Partially Observable Markov Decision Process (POMDP) with $N$ \textbf{interacting} agents (Littman, 1994). This formulation assumes a set of state, $\mathcal{S}$, containing all the states characterizing the environment; a set of actions $\mathcal{A} =\left\{\mathcal{A}_{1},\dots,\mathcal{A}_{N} \right\}$ where each $ \mathcal{A}_{i} $ is a set of possible actions for the $ i^{th} $ agent; and a set of observations  $\mathcal{O} = \left\{ \mathcal{O}_{1}, \dots ,\mathcal{O}_{N} \right\}$, one for each agent in the environment. Each $ \textbf{o}_{i}$ in $\mathcal{O}_{i}$ indicates a partial characterization of the current state and is private for the agent. State transitions are controlled by the current state and one action from each agent: $\mathcal{T:S \times}\mathcal{A}_{1} \times \ldots \times \mathcal{A}_{N} \mapsto \mathcal{S}$, $ i = 1, \ldots, N $. Each agent also has an associated reward function based on the joint actions, \({r}_{i}\mathcal{:S \times}\mathcal{A}_{1} \times \ldots \times \mathcal{A}_{N}\mapsto{\mathbb{R}}\), \( i = 1, \ldots, N \). The joint policy is given by \( \pi := (\pi_{\theta_1}, \ldots, \pi_{\theta_N}) \subset \Pi\). Because the rewards of the agents depend on the joint policy, the return for each agent is denoted as follows:

\begin{equation}\label{Eq1}
	R_{i}^\pi (\mathcal{S}) = E(\sum_{k=0}^{\infty } {\gamma }^{k}r_{i,k+1}|s_0\in \mathcal{S}, \pi)
\end{equation}
where $\gamma$ is discount factor of future rewards, $s_0$ is the initial state.

\subsection{Q-learning Algorithms}\label{qlearning}
Q-learning is a classical RL method that fits the Q-function with a parametric model $Q_{\theta}$. The action-value function in Q-learning for policy $\bm{\pi}$ is given as $Q^{\bm{\pi}}\left(s,a \right)\mathbb{= E\lbrack}R|s^{t} = s,\ a^{t} = a\rbrack$. This Q function can be recursively rewritten as $Q^{\bm{\pi}}\left( s,a \right) = \mathbb{E}_{s^{'}}\left\lbrack r\left( s,a \right) + \gamma\mathbb{E}_{a^{'}\sim\bm{\pi}}\left\lbrack Q^{\bm{\pi}}\left( s^{'},a^{'} \right) \right\rbrack \right\rbrack$. On the other hand, deep Q-networks (DQN) learns the action-value function $Q^{*}$ corresponding to the optimal policy by minimizing the loss:

\begin{equation}\label{Eq2}
     \mathcal{L}\left( \theta \right) = \mathbb{E}_{s,a,r,s^{'}}\left\lbrack \left( Q^{*}\left( s,a \middle| \theta \right) - y \right)^{2} \right\rbrack, \text{where}\  y = r + \gamma\max_{a^{'}}{{\overline{Q}}^{*}\left( s^{'},a^{'} \right)}
\end{equation}
where $\overline{Q}$ is a target Q function whose parameters are intermittently updated with the latest $\theta$, which helps to stabilize learning.

\subsection{Policy Gradient Algorithms}\label{ddpg}
The major idea of policy gradient (PG) algorithms is to directly adjust the parameters $\theta$ of the policy in order to maximize the objective $ J\left( \theta \right) = \mathbb{E}_{s\sim p^{\bm{\pi}},\ a\sim\bm{\pi}_{\theta}}\left\lbrack R \right\rbrack$ by taking steps in the direction of \(\nabla_{\theta}J\left( \theta \right)\). Using the Q function defined previously, the gradient of the policy can be written as:

\begin{equation}\label{Eq3}
	{\nabla_{\theta}J\left( \theta \right)\mathbb{= E}}_{s\sim p^{\bm{\pi}},\ a\sim\bm{\pi}_{\theta}}\lbrack\nabla_{\theta}\log{\bm{\pi}_{\theta}\left( a \middle| s \right)Q^{\bm{\pi}}\left( s,a \right)\rbrack}
\end{equation}
where $p^{\bm{\pi}}$ is the state distribution. There are several methods to estimate $Q^{\bm{\pi}}$, resulting in different practical algorithms, for instance, REINFORCE \cite{Duan2016}, actor-critic \cite{Konda2000}, etc. Specifically, the actor-critic algorithm uses an approximation, $Q_{\varpi}(s,a)$, to estimate the true action-value function $Q^{\bm{\pi}}\left(s,a \right)$.

A recent extension of the policy gradient framework using deterministic policies \(\mu_{\theta}\mathcal{:S \mapsto A}\), named deterministic policy gradient (DPG), combined the benefits of DQN and actor-critic algorithms \cite{Silver2014}. In particular, if the action space \(\mathcal{A}\) and the policy \(\bm{\mu}\) are continuous (then the Q-function is presumed to be differentiable with respect to the action argument), the gradient of the objective
\(J\left( \theta \right) = \mathbb{E}_{s\sim p^{\bm{\mu}}}\left\lbrack R\left( s,a \right) \right\rbrack\) can be rewritten as:

\begin{equation}\label{Eq4}
	\nabla_{\theta}J\left( \theta \right) = \mathbb{E}_{s\mathcal{\sim D}}\left\lbrack \nabla_{\theta}\bm{\mu}_{\theta}\left( a \middle| s \right)\nabla_{a}Q^{\bm{\mu}}\left( s,a \right)\left. \  \right|_{a = \mu_{\theta}(s)} \right\rbrack
\end{equation}

One of the major advantages of DPGs is that although stochastic policy gradients are integrated into both state and action spaces, DPGs are only integrated over the state space, requiring fewer samples in problems with large action spaces. Previous studies have shown that DPG greatly improves the stochastic policy gradient equivalence in high-dimensional continuous control problems \cite{Silver2014}.

\subsection{Multi-Agent Deep Deterministic Policy Gradient Algorithm}\label{maddpg}
In the article of \cite{Lowe2017}, the authors proposed a multi-agent extension of the actor-critic policy gradient methods where the critic is augmented with extra information about the policies of other agents. More specifically, in the game with $N$ agents, let $\bm{\mu}_{i}$ be the set of continuous policies with respect to parameters $\bm{\theta}_{i}$. Then, the gradient of the expected return for agent $i$ is equal to
\begin{equation}\label{Eq5}
\nabla_{\theta_{i}}J\left( \mu_{i} \right) = \ \mathbb{E}_{x,a\mathcal{\sim D}}\left\lbrack {\nabla_{\theta}}_{i}\bm{\mu}_{i}\left( a_{i}|o_{i} \right)\nabla_{a_{i}}Q_{i}^{\bm{\mu}}\left( x,a_{1},\ldots,\ a_{N} \right)|_{a_{i} = \bm{\mu}_{i}(o_{i})} \right\rbrack
\end{equation}
where \(Q_{i}^{\bm{\mu}}\left( x,a_{1},\ldots,\ a_{N} \right)\) is the \emph{centralized action-value function} that takes as input of all agents and state information \(x\), and outputs the Q-value for agent \(i\). \(\mathcal{D}\) is the experience reply buffer and contains the element of \(\left( x,\ x^{'},\ a_{1},\ldots,\ a_{N},r_{1},\ldots,\ r_{N} \right)\), which keeps a memory of previous action-reward pairs and train the network with samples. Furthermore, \(Q_{i}^{\bm{\mu}}\) is updated as:
\begin{equation}\label{Eq6}
	\mathcal{L}\left( \theta_{i} \right) = \mathbb{E}_{\textbf{x},a,r,\textbf{x}^{'}\mathcal{\sim D}}\left\lbrack \left( Q_{i}^{\bm{\mu}}\left( x,a_{1},\ldots,\ a_{N} \right)-y \right)^{2} \right\rbrack
\end{equation}
where $y$ is the expected return computed by target network, and is equals to
\begin{equation}\label{Eq7}
	y = r_{i} + \gamma Q_{i}^{\bm{\mu}^{'}}\left( \textbf{x}^{'},a_{1}^{'},\ldots,\ a_{N}^{'} \right)|_{a_{j}^{'} = \bm{\mu}_{j}^{'}(o_{j})}
\end{equation}
where $\bm{\mu}^{'} = \left\{ \bm{\mu}_{\theta_{1}^{'}},\ \ldots,\bm{\mu}_{\theta_{N}^{'}} \right\}$ is the set of \emph{target policies} with delayed parameters $\theta_{i}^{'}$.

\section{Methodology}
\subsection{Problem Statement}
We consider a road network with $N\left( N > 1 \right)$ signalized intersections, and adopt a multi-agent extension of partially observable Markov decision process characterized by the components $\left\langle \mathcal{S,O,A,P},r,\pi,\gamma \right\rangle$ as mentioned in Section \ref{MDP}. More specifically, we assume that each intersection is controlled with a control unit that is able to adapt its phase setting based on the traffic patterns or human inputs. Traffic sensors are implemented at each lane of each intersection and are able to automatically collect real-time traffic data and send them to the nearest control unit. A novel knowledge-sharing mechanism enabled reinforcement learning approach is proposed to solve the network-wide ATSC optimization problem. Through the proposed knowledge-sharing framework, each agent is able to explicitly capture relevant traffic state of the system observed by other agents (i.e., control units), and therefore could simultaneously learn from both its private observations and the collectively learned representation of the system that accumulates through experiences coming from others to optimize its policy. The agent can also interpret the shared knowledge in its own unique way as needed to optimize its policy. In addition, this knowledge-sharing mechanism is designed to be used in both training and execution.

\subsection{Knowledge Sharing Architecture}
\begin{figure}[H]
	\centering
	\includegraphics[width=1\linewidth]{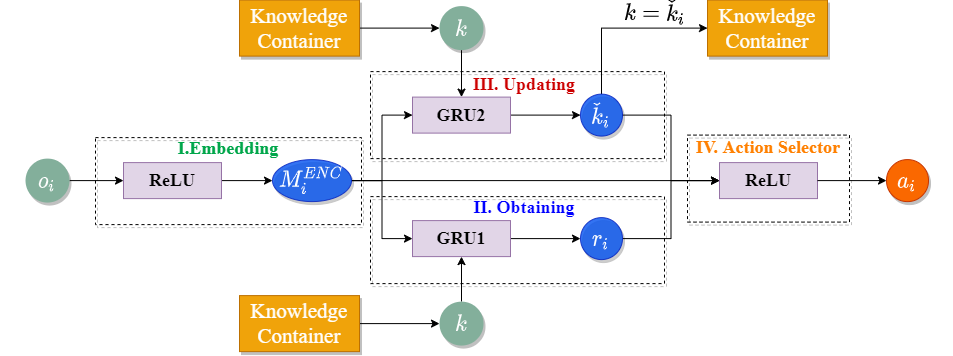}
	\caption{Flowchart of knowledge-sharing}
	\label{fig:operation}
\end{figure}

As aforementioned,  a knowledge-sharing mechanism is proposed as the communication protocol for the agent to share and utilize its understanding of its and others' observations.  A centralized \textit{ knowledge container} $\mathcal{C}$ is designed to store the \textit{collective knowledge} $\textbf{k} \in\mathbb{R}^{1 \times K}$ that progressively collected by all agents as they interact. $\textbf{k}$ is a highly compact vector and represents the collective understanding of the environment by all agents. Each agent has its own way to interpret it and rebuild it.  When making a decision, the agents need to taking both its private observation but also its own comprehension of the collective knowledge $\textbf{k}$ into account via the help of the communication protocol.  As presented in Fig. \ref{fig:operation},  each agent needs to first execute an \textit{Observation Embedding} operation to encode its private observation into a latent space. Then, prior to taking an action, the agent has to execute an \textit{Knowledge Obtaining} operation in order to access the knowledge container, and initially retrieve and interpret the knowledge stored in it. After obtaining the knowledge, the agent will perform a \textit{Knowledge Updating} operation, and update the current knowledge content based on its own observations. During training, these operations are learned without imposing any prior constraints on the nature of the collective knowledge. Whilst in the execution process, the agents use the communication protocol that they have learned to obtain and update the knowledge over the whole episode. In order to build a trainable \textit{end-to-end} model, we use deep neural networks as function approximators for policies, and use learnable gated functions to facilitate each agent's interactions with the knowledge.  The details of these operations are introduced and discussed below.  In order to make the description more concise, time-related subscripts in the formulas will be omitted once there is no ambiguity.

\noindent \textbf{I. Observation Embedding.}
At time $t$, the up-to-date $n$-dimensional private observations received by agent $i$, i.e., the traffic volume on each entrance lane and the current phase at the $i$-th intersection, $\textbf{o}_{i}$, are mapped on to an embedding latent space $\textbf{M}^{ENC}_{i} \in \mathbb{R}^{1 \times m}$, through a multi-layer perceptron:
\begin{equation}\label{Eq8}
	\textbf{M}^{ENC}_{i}=\text{Embed}(\textbf{o}_{i})=\text{ReLU}(\textbf{o}_{i}\textbf{W}_{i,oM}+\textbf{b}_{i,M})
\end{equation}
where $\textbf{W}_{i,oM} \in \mathbb{R}^{n \times m}$ and $\textbf{b}_{i,M} \in \mathbb{R}^{1 \times m}$ are the	weight matrix and the bias vector to learn, and ReLU represents the ReLU activation function. The second element of the subscript of $\textbf{W}_{i,oM}$ indicates that the vector is related to the observation $\textbf{o}$ and the embedding result $\textbf{M}^{ENC}$. Similarly, the $M$ in the subscript of $\textbf{b}_{i,M}$ implies this vector is related to $\textbf{M}^{ENC}$. The following formulas will also use this notation method. The embedding $\textbf{M}^{ENC}_{i}$ represents latent state of the current traffic condition at the $i$-th intersection, and plays a fundamental role in selecting new actions and in the following knowledge obtaining and updating phases.

\noindent \textbf{II. Knowledge Obtaining.}
This obtaining operation is performed after the encoding operation and allows the agent to acquire and comprehend the collective knowledge $\textbf{k}$ previously captured by other agents. By interpreting this knowledge content, the agent has access to what others have learned. The intuition behind this setting is that for the intersections in the network, some of them may have strong correlations, i.e., some traffic patterns always travel through them in a sequence; while others have limited interactions since only a few travelers are likely to choose the routes that consist of them conjointly. Therefore, taking observations obtained by other agents into account may have significant favorable impacts on the agent's control performance when making decisions. Instead of gathering all historical observations of others,  we use an \textit{update gate and reset gate} mechanism inspired by Gated Recurrent Unit (GRU) \cite{Chung2014} to let the agent itself decides what data in the collective knowledge is kept or threw away. Therefore, compared to fully communicating, this operation could remarkably enhance the execution efficiency.

Let the vector $\textbf{r}_{i} \in \mathbb{R}^{1 \times K}$ denote the knowledge content that the agent $i$ decides to obtain at time $t$, then the update gate vector $\textbf{z}_{i} \in \mathbb{R}^{1 \times K}$ that helps the model determine how much of the past knowledge (from previous time steps) needs to be passed along to the future can be written as
\begin{equation}\label{Eq9}
	\textbf{z}_{i} = \sigma_{g}(\textbf{M}^{ENC}_{i}\textbf{W}_{i,Mz}+\textbf{k}\textbf{W}_{i,kz}+\textbf{b}_{i,z})
\end{equation}
where $\textbf{k}$ is the knowledge content already stored in the container, $\textbf{W}_{i,Mz} \in \mathbb{R}^{m \times K}$ and $\textbf{W}_{i,kz} \in \mathbb{R}^{K \times K}$ are the weights to be learned, $\textbf{b}_{i,z} \in \mathbb{R}^{1 \times K}$ is the bias term, and $\sigma_{g}$ is the sigmoid activation function to squash the result between 0 and 1.

Essentially, the reset gate vector $\textbf{l}_{i} \in \mathbb{R}^{1 \times K}$ is used from the model to decide how much past knowledge to forget. To calculate it, we use a similar formula like that in Eq. \ref{Eq9} :
\begin{equation}\label{Eq10}
	\textbf{l}_{i} = \sigma_{g}(\textbf{M}^{ENC}_{i}\textbf{W}_{i,Ml}+\textbf{k}\textbf{W}_{i,kl}+\textbf{b}_{i,l})
\end{equation}
where $\textbf{W}_{i,Ml} \in \mathbb{R}^{m \times K}$ and $\textbf{W}_{i,kl} \in \mathbb{R}^{K \times K}$ are also learnable weight matrices, and $\textbf{b}_{i,l} \in \mathbb{R}^{1 \times K}$ is the bias term. Assuming $\textbf{\~{r}}_{i} \in \mathbb{R}^{1 \times K}$ is the candidate knowledge obtaining vector, and its calculation method is as follows:
\begin{equation}\label{Eq11}
	\textbf{\~{r}}_{i}=\text{tanh}(\mathbf{M}^{ENC}_{i}\mathbf{W}_{i,Mk}+\mathbf{l}_{i}\odot(\mathbf{k}\mathbf{W}_{i,kk})+\mathbf{b}_{i,k})
\end{equation}
where $\odot$ is the Hadamard (element-wise) product operator between two vectors, tanh is the activation function and is applied to every element of its input. Here we use a nonlinearity in the form of tanh to ensure that the values in $\textbf{\~{r}}_{i}$ remain in the interval $(-1,1)$. The result is a candidate since we still need to consider the action of the update gate. Therefore, at the last step, we incorporate the effect of the update gate $\mathbf{z}_{i}$. This determines the extent to which the final knowledge content ($\mathbf{r}_{i}$) obtained by agent $i$ at time $t$ is just that already exists in the container($\mathbf{k}$) and by how much the new candidate knowledge ($\textbf{\~{r}}_{i}$) is used. The update gate $\mathbf{z}_{i}$ can be used for this purpose, simply by taking elementwise convex combinations between $\textbf{\~{r}}_{i}$ and $\mathbf{z}_{i}$, that is,
\begin{equation}\label{Eq12}
	\mathbf{r}_{i}=\mathbf{z}_{i}\odot\mathbf{k} +(1-\mathbf{z}_{i})\odot \textbf{\~{r}}_{i}
\end{equation}

\begin{figure}[h]
	\centering
	\includegraphics[width=1\linewidth]{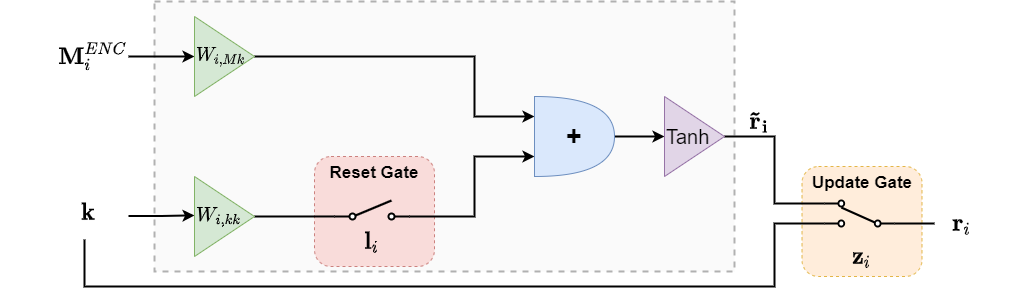}
	\caption{A Brief Summary of Knowledge Obtaining Operation}
	\label{fig:knowledgeobtain}
\end{figure}

The knowledge obtaining operation can be further summarized as Fig. \ref{fig:knowledgeobtain}. We use a single pole single throw switch to represent the effect of the reset gate, and a single pole double throw switch to demonstrate the effect of the update gate. When $\mathbf{l}_{i}$ equals to 0, indicating there are no impacts of previous knowledge on the current knowledge content; otherwise, the impacts exist. It should also be noteworthy that when $\mathbf{z}_{i}$ equals to 0 and $\mathbf{l}_{i}$ equals to 1, this operation degenerates into a standard RNN. In summary, the reset and update gates help the agent capture short-term and long-term dependencies in sequences respectively. All the weights and bias vectors in this operation are learnable and agent-specific, indicating that each agent is able to interpret the knowledge in its own unique way.

\noindent \textbf{III. Knowledge Updating.}
In the updating phase, we also use a GRU based operation to let the agent itself decides what in its observations should be updated and what to discarded. Assuming the agent first generates a candidate knowledge updating content, $\mathbf{\widehat{k}}_{i} \in \mathbb{R}^{1 \times K}$, which depends on its own observations and the stored knowledge through a non-linear mapping, that is,
\begin{equation}\label{Eq13}
	\mathbf{\widehat{k}}_{i} = \text{tanh}(\mathbf{M}^{ENC}_{i}\mathbf{W}_{i,M\widehat{k}}+\mathbf{p}_{i}\odot(\mathbf{k}\mathbf{W}_{i,k\widehat{k}})+\mathbf{b}_{i,\widehat{k}})
\end{equation}
where $\mathbf{W}_{i,M\widehat{k}} \in \mathbb{R}^{m \times K}$ and $\mathbf{W}_{i,k\widehat{k}} \in \mathbb{R}^{K \times K}$ are weights to learn, $\mathbf{b}_{i,\widehat{k}}  \in \mathbb{R}^{1 \times K}$ is the bias vector, and $\mathbf{p}_{i} \in \mathbb{R}^{1 \times K}$ is the reset gate which is similar to Eq. \ref{Eq10}, and can be further written as
\begin{equation}\label{Eq14}
	\mathbf{p}_{i} = \sigma_{g}(\mathbf{M}^{ENC}_{i} \mathbf{W}_{i,Mp} +\mathbf{k}\mathbf{W}_{i,kp}+\mathbf{b}_{i,p})
\end{equation}
where $\mathbf{W}_{i,Mp} \in \mathbb{R}^{m \times K}$ and $\mathbf{W}_{i,kp} \in \mathbb{R}^{K \times K}$ are learnable weight matrices, and  $\mathbf{b}_{i,\widehat{k}}  \in \mathbb{R}^{1 \times K}$ is the bias vector.

Following Eq. \ref{Eq9}, we can have the update gate  $\mathbf{q}_{i} \in \mathbb{R}^{1 \times K}$ as
\begin{equation}\label{Eq15}
	\mathbf{q}_{i} = \sigma_{g}(\mathbf{M}^{ENC}_{i}\mathbf{W}_{i,Mq}+\mathbf{k}\mathbf{W}_{i,kq}+\mathbf{b}_{i,q})
\end{equation}
where $\mathbf{W}_{i,Mq}  \in \mathbb{R}^{m \times K}$, $\mathbf{W}_{i,kq} \in \mathbb{R}^{K \times K}$ and $\mathbf{b}_{i,q} \in \mathbb{R}^{1 \times K}$ are learnable weight metrics and bias vector, respectively.

Agent $i$ then finally generates an updated knowledge $\mathbf{\check{k}}_{i}$ as a weighted combination based on its current observations and previous knowledge $\mathbf{k}$, as follows:
\begin{equation}\label{Eq16}
	\mathbf{\check{k}}_{i}= \mathbf{q}_{i}\odot\mathbf{k} +(1-\mathbf{q}_{i})\odot\mathbf{\widehat{k}}_{i}
\end{equation}
The updated knowledge $\mathbf{\check{k}}_{i}$ is stored in the container $\mathcal{C}$ as $\mathbf{k}$ and made accessible to all the agents. At each time step, agents sequentially obtain and update the content of knowledge using above procedures.

\noindent \textbf{IV. Action Selector.} Once the agent completing both obtaining and updating operations, it can develop its action, $a_{i}$ which depends on its current encoding of its observations $\mathbf{M}^{ENC}_{i}$, its own interpretation of the current knowledge content $\mathbf{r}_{i}$ and its updated version $\mathbf{\check{k}}_{i}$, that is
\begin{equation}\label{Eq17}
	 {a}_{i}=\phi_{\theta_{i}^{a}}^{ACT}(\mathbf{M}^{ENC}_{i}, \mathbf{r}_{i},\mathbf{\check{k}}_{i})
\end{equation}
where $\phi_{\theta_{i}^{a}}^{ACT}$ is a neural network parameterized by $\theta_{i}^{a}$. Therefore, the resulting policy function can be further written as a composition of functions:
\begin{equation}\label{Eq18}
	\bm{\mu}_{\theta_{i}}(\mathbf{o}_{i},\mathbf{k})= \varphi_{\theta_{i}^{a}}^{ACT} \left( \textbf{M}^{ENC}_{i} (\mathbf{o}_{i}) ,\mathbf{r}_{i} (\mathbf{o}_{i},\mathbf{k}),\mathbf{\check{k}}_{i} (\mathbf{o}_{i},\mathbf{k})\right)
\end{equation}
in which $\theta_{i}$ contains all the relevant parameters from the encoding, obtaining, and updating phases.

\begin{figure}[H]
	\centering
	\includegraphics[width=0.9\linewidth]{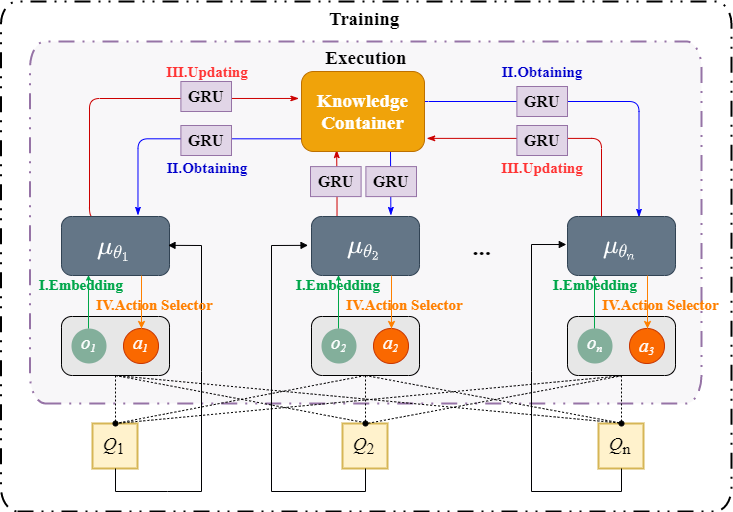}
	\caption{Framework of KS-DDPG algorithm}
	\label{fig:framework}
\end{figure}

\subsection{Learning Algorithm}
All agent-specific policy parameters, i.e.,  $\theta_{i}$, are learned end-to-end. Inspired by the MADDPG algorithm introduced in Section \ref{maddpg}, an actor-critic model enables knowledge sharing within a centralized learning and decentralized execution framework is adopted. The framework of the proposed KS-DDPG algorithm  is presented in Fig. \ref{fig:framework}. In actor-critic models, the actor is used to select actions, while the critic is used to evaluate the actor's actions and provide feedback. Since the input of the policy of our model (as described in Eq. \ref{Eq18}) is $(\mathbf{o}_{i},\mathbf{k})$, we therefore use Eq. \ref{Eq5} to rewrite the gradient of the expected return for agent $i$ as
\begin{equation}\label{Eq19}
	\nabla_{\theta_{i}}J(\bm{\mu}_{\theta_{i}})=\mathbb{E}_{\mathbf{x},a,\mathbf{k}\sim\mathcal{D}}\left[\nabla_{\theta_{i}}\bm{\mu}_{\theta_{i}}(\mathbf{o}_{i},\mathbf{k})\nabla_{a_{i}}Q^{\bm{\mu}_{\theta_{i}}}(\mathbf{x}, a_{1},\ldots,a_{N})|_{a_{i}=\bm{\mu}_{\theta_{i}}(\mathbf{o}_{i},\mathbf{k})} \right]
\end{equation}
where $\mathcal{D}$ is the replay buffer storing past experiences. It contains transitions in the form of $(\mathbf{x},\mathbf{x}^{'}, a_{1},\ldots,a_{N}, \mathbf{k},r_{1},\ldots,r_{N})$. $Q^{\bm{\mu}_{\theta_{i}}}$ is updated according to Eqs. \ref{Eq6} and \ref{Eq7}. The detailed pseudo-code of the learning and execution procedure is presented in Algorithm \ref{Algorithm1}.

\begin{breakablealgorithm}
	\caption{KS-DDPG Algorithm for $N$ Interacting Agents}
	\label{Algorithm1}
	\begin{algorithmic}[1]
		\STATE Initialize actors $(\bm{\mu}_{\theta_{1}},\dots,\bm{\mu}_{\theta_{N}})$ and critics networks $(Q_{\theta_{1}},\dots,Q_{\theta_{N}})$
		\STATE Initialize actor target networks $(\bm{\mu}^{'}_{\theta_{1}},\dots,\bm{\mu}^{'}_{\theta_{N}})$ and critic target networks $(Q_{\theta_{1}}^{'},\dots,Q_{\theta_{N}}^{'})$
		\STATE Initialize replay buffer $\mathcal{D}$
		\FOR {episode=1 to $E$}
		\STATE Initialize a random process $\mathcal{N}$ for action exploration
		\STATE Initialize knowledge container $\mathcal{C}$, and receive initial state $\mathbf{x}$
		\FOR {$t=1$ to \textbf{max episode length}}
		\FOR {agent $i=1$ to $N$}
		\STATE Receive observation $\mathbf{o}_{i}$ and the collective knowledge $\mathbf{k}$
		\STATE Set $\mathbf{k}_{i}=\mathbf{k}$
		\STATE Generate knowledge encoding vector $\mathbf{M}^{ENC}_{i}$ using Eq. \ref{Eq8}
		\STATE Generate knowledge obtaining vector $\mathbf{r}_{i}$ using Eq. \ref{Eq12}
		\STATE Generate knowledge updating vector $\mathbf{\check{k}}_{i}$ using Eq. \ref{Eq16}
		\STATE Store the new knowledge in the container $\mathbf{k}\leftarrow \mathbf{\check{k}}_{i}$
		\STATE Generate new time dependent noise instance $\mathcal{N}_{t}$
		\STATE Select action ${a}_{i}$ using Eq. \ref{Eq17}
		\ENDFOR
		\STATE Set $\mathbf{x}=(\mathbf{o}_{1},\ldots,\mathbf{o}_{N})$ and $\bm{\Phi}=(\mathbf{k}_{1},\ldots,\mathbf{k}_{N})$
		\STATE Execute actions $\mathbf{a}=(a_{1},\ldots,a_{N})$, observe rewards $r$ and next state $\mathbf{x}^{'}$
		\STATE Store $(\mathbf{x},\mathbf{x}^{'},\mathbf{a},\bm{\Phi},r)$ in replay buffer $\mathcal{D}$
		\ENDFOR
		\FOR {agent $i=1$ to $N$}
		\STATE Sample a random minibatch $\Theta$ of $S$ samples $(\mathbf{x},\mathbf{x}^{'},\mathbf{a},\bm{\Phi},r)$ from $\mathcal{D}$
		\STATE Set $y=r_i+\gamma Q^{\bm{\mu}^{'}_{\theta_{i}}}(\mathbf{x}^{'}, a_{1}^{'},\ldots,a_{N}^{'})|_{a_{j}^{'}=\bm{\mu}^{'}_{\theta_{j}}(\mathbf{o}_{j},\mathbf{k}_{j})}$
		\STATE Update critic by minimizing the loss:
		\begin{equation*}
			\mathcal{L}(\theta_{i})=\dfrac{1}{S}\sum\limits_{(\mathbf{x},\mathbf{x},\mathbf{a},\bm{\Phi},r)\in \Theta} \left( y-Q^{\bm{\mu}_{\theta_{i}}}(\mathbf{x}, a_{1},\ldots,a_{N})\right) ^{2}
		\end{equation*}
		\STATE Update actor using the sampled policy gradient:
		\begin{equation*}
			\nabla_{\theta_{i}}J\approx \dfrac{1}{S}\sum\limits_{(\mathbf{x},\mathbf{x}^{'},\mathbf{a},\bm{\Phi},r)\in \Theta} \left( \nabla_{\theta_{i}}\bm{\mu}_{\theta_{i}}(\mathbf{o}_{i},\mathbf{k})\nabla_{a_{i}}Q_{i}^{\bm{\mu}_{\theta_{i}}}(\mathbf{x}, a_{1},\ldots,a_{i},\ldots,a_{N})|_{a_{i}=\bm{\mu}_{\theta_{i}}(\mathbf{o}_{i},\mathbf{k}_{i})}\right)
		\end{equation*}
		\ENDFOR
		\STATE Update target network parameters for each agent $i$:
		\begin{equation*}
			\theta_{i}^{'}=\tau\theta_{i}+(1-\tau)\theta_{i}^{'}
		\end{equation*}
		\ENDFOR

	\end{algorithmic}
\end{breakablealgorithm}

As illustrated in Algorithm \ref{Algorithm1}, only the learned actors are used to make decisions and select actions during execution. Each agent takes its action successively.  The agent receives its private observations, obtains the knowledge stores in $\mathcal{C}$, generates a new version of the knowledge, stores it back into $\mathcal{C}$ and selects its action $a_i$ using $\bm{\mu}_{\theta_{i}}$. The policy of the next agent is then driven by the update $\mathcal{C}$. An experience replay buffer contains the tuples $(\mathbf{x},\mathbf{x}^{'},\mathbf{a},\mathbf{\Phi},r)$ is adpoted to record experiences of all agents. By minimizing the loss function $\mathcal{L}(\theta_{i})$, the model attempts to improve the estimate of the critic network which is used to improve the policy itself through update with the sampled policy gradient $\nabla_{\theta_{i}}J$.

\subsection{Agent Design}
\noindent \textbf{State Space and Observation Space.}
We assume that at time $t$ agent $i$ can only observe part of the system state $\textbf{s}_{i}\in\mathcal{S}$ as its observation $\textbf{o}_{i}\in\mathcal{O}$. Although sensing additional combinations of data from upstream and downstream may have more advantageous impacts on the control performance, the dramatically increased state space can readily result in an intractability model. In order to limit the state space, two effortlessly collected elements, including the current phase and traffic volume in each entrance lane, were selected as key indicators to describe the state of the agent. The volume of a lane is defined as the number of vehicles on the lane, which equals the sum of queuing vehicles and moving vehicles on the entrance lane.

\noindent \textbf{Set of Actions.}
Different action schemes have significant influences on traffic signal control strategies. In current ATSC systems, such as SCOOT and SCATS, there are several different control strategies, for instance, phase shift, phase length change, phase itself, etc., or a combination of some of the above. Undoubtedly, using a flexible candidate action set could considerably increase the action space as well as the computational burden. In contrast, fixing the phase structure and phase sequence during the optimization process may bring forth lags in timing plan updating, making it unable to respond to the traffic flow fluctuation. Therefore, this kind of control method is a responsive ATSC rather than a real-time ATSC.

\begin{figure}[H]
	\centering
	\includegraphics[width=1\linewidth]{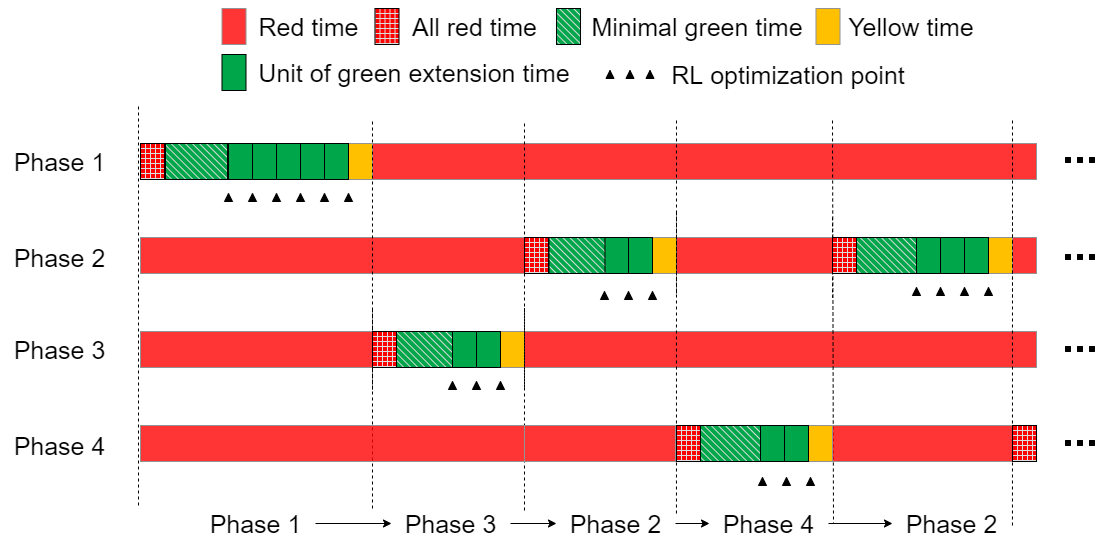}
	\caption{Diagram of Candidate Actions in an Acyclic ATSC}
	\label{fig:rlconrtol}
\end{figure}

In order to seek a balance between the benefits of flexible action setting and its impacts on the ease of problem handling, it is assumed that the cycle lengths are not fixed in this case, but can only be adjusted between the maximum and minimum limits to ensure practicality. As shown in Fig. \ref{fig:rlconrtol}, the agent accounts for a variable phasing sequence in an acyclic timing scheme with an unfixed cycle length and an unsettled phasing sequence. More specifically, the control action is either to extend the current phase (i.e., choose the number of units of green extension) or to switch to any other phase, and consequently, some unnecessary phases may be skipped according to the fluctuations in traffic. In addition, each chosen phase is assumed to have a length limit with predetermined minimal green time, yellow time, and red time as the settings in engineering practice. It can be considered that it is desirable to limit the variability of consecutive cycles and ensure traffic safety, although some degradation of performance is likely.

\noindent \textbf{Reward.}
Due to the complex spatial-temporal structure of traffic data, the reward should be spatially decomposable and can be easily measured after an action. Therefore, the reward of an agent is defined as the reduction of the average delay of vehicles at all entrance lanes associated with the control units, i.e., the difference between the average delays of two successive decision points. A positive value of the reward indicates that the delay is reduced by this value after executing the selected action; otherwise, this action results in an increase in the average delay.

\section{Experiments}

\subsection{General Setups}

\begin{figure}[H]
	\centering
	\includegraphics[width=1\linewidth]{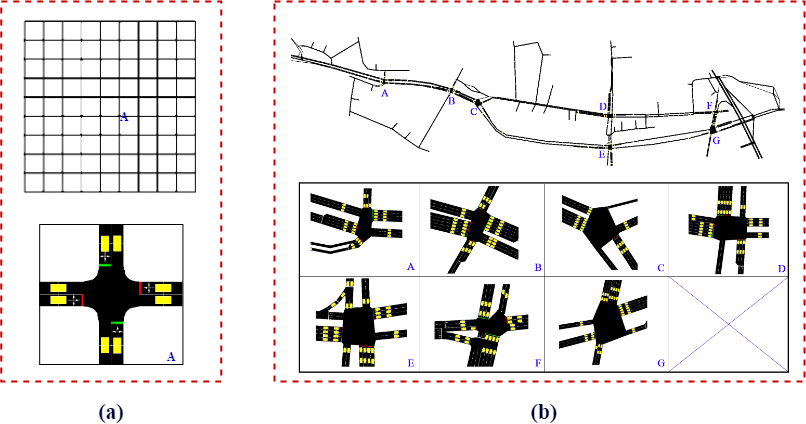}
	\caption{Network Structures in Numerical Experiments (a)\textit{Grid} (b) \textit{MoCo}}
	\label{fig:network}
\end{figure}

The proposed KS-DDPG algorithm is evaluated in the Simulation of Urban MObility (SUMO) platform. SUMO is an open-source traffic simulation package designed by the Institute of Transportation Systems at the German Aerospace Center to handle large road networks and has already been widely used in transportation-related research \cite{Krajzewicz2012}. SUMO was chosen because of its scalability as it includes an API called TraCI (Traffic Control Interface). TraCI allows users to query and modify the simulation status and extend the existing SUMO functionality. Finally, the OpenAI Gym which is a framework designed for the development and evaluation of reinforcement learning algorithms runs compatible with SUMO to set up a simulation environment \cite{Brockman2016}.

The numerical experiments are carried out through two different networks: (1) \textit{Grid}: a Manhattan-style virtual 10$\times$10 grid network with the same signalized intersection structures; and (2) \textit{MoCo}: a real-world network extracted from downtown Montgomery County, Maryland with heterogeneous intersections of both signalized and non-signalized ones. The detailed structures of the two networks are illustrated in Fig. \ref{fig:network}. In addition, Table \ref{movement} presents the detailed information of lanes and phase combinations at each signalized intersection in both experiments.

\begin{table}[H]
	\centering
	\caption{Information regarding signals under the control}
	\label{movement}
	\resizebox{\textwidth}{!}{%
		\begin{tabular}{|l|l|l|l|l|}
			\hline
			Experiment &
			Intersection ID &
			Lanes of Approaches &
			Legal Movement of Phases &
			Aberrations \\ \hline
			Grid &
			All &
			\begin{tabular}[c]{@{}l@{}}NB:1*LTR \\ SB:1*LTR\\ WB:1*LTR\\ EB:1*LTR\end{tabular} &
			\begin{tabular}[c]{@{}l@{}}NL+NT\\ SL+ST\\ WL+WT\\ EL+ET\end{tabular} &
			\multirow{8}{*}{\begin{tabular}[c]{@{}l@{}}Approaches: \\ NB: Northbound\\ SB: Southbound\\ WB: Westbound \\ EB: Eastbound\\ \\ Movement of a Lane:\\ T: Through\\ L: Left turning\\ R: Right turning\\  \\ Legal Movement of a Phase:\\ First digit: Approach \\ Second digit:\\ Turning Movement\end{tabular}} \\ \cline{1-4}
			\multirow{7}{*}{MoCo} &
			A &
			\begin{tabular}[c]{@{}l@{}}NB:1*L+1*TR\\ SB:1*L+1*TR\\ WB:1*L+2*T+1*TR \\ EB:1*L+2*T+1*TR\end{tabular} &
			\begin{tabular}[c]{@{}l@{}}NL+NT \\ SL+ST\\ WL+WT\\ EL+ET\end{tabular} &
			\\ \cline{2-4}
			&
			B &
			\begin{tabular}[c]{@{}l@{}}NB:1*LT+1*R\\ SB:1*LT+1*R\\ WB:1*LT+3*T+1*R\\ EB:1*L+3*T+1*R\end{tabular} &
			\begin{tabular}[c]{@{}l@{}}NL+NT\\ SL+ST\\ WL+WT\\ EL+ET\end{tabular} &
			\\ \cline{2-4}
			&
			C &
			\begin{tabular}[c]{@{}l@{}}NB:1*T\\ WB:3*L+2*T\\ EB:2*T+1*R\end{tabular} &
			\begin{tabular}[c]{@{}l@{}}NT\\ WL+WT\\ ET\end{tabular} &
			\\ \cline{2-4}
			&
			D &
			\begin{tabular}[c]{@{}l@{}}NB:1*L+2*T+2*R\\ SB:1*L+1*T+1*TR\\ WB:2*L+2*T+1*R\\ EB:1*L+1*T+1*TR\end{tabular} &
			\begin{tabular}[c]{@{}l@{}}NL+NT\\ SL+ST\\ WL+WT\\ EL+ET\end{tabular} &
			\\ \cline{2-4}
			&
			E &
			\begin{tabular}[c]{@{}l@{}}NB:1*L+1*TR\\ SB:1*L+2*T+1*TR\\ WB:1*L+2*T+1*R\\ EB:1*LT+1*T+1*TR\end{tabular} &
			\begin{tabular}[c]{@{}l@{}}NL+NT\\ SL+ST\\ WL+WT\\ EL+ET\end{tabular} &
			\\ \cline{2-4}
			&
			F &
			\begin{tabular}[c]{@{}l@{}}NB:1*L+1*T+1*TR\\ SB:2*L+1*T+1*TR\\ WB:1*LT\\ EB:1*L+1*TR\end{tabular} &
			\begin{tabular}[c]{@{}l@{}}NL+NT\\ SL+ST\\ WL+WT\\ EL+ET\end{tabular} &
			\\ \cline{2-4}
			&
			G &
			\begin{tabular}[c]{@{}l@{}}NB:2*L+1*TR\\ SB:1*LT+1*TR\\ WB:1*L+1*T+1*TR\\ EB:1*L+2*T+2*R\end{tabular} &
			\begin{tabular}[c]{@{}l@{}}NL+NT\\ SL+ST\\ WL+WT\\ EL+ET\end{tabular} &
			\\ \hline
		\end{tabular}%
	}
\end{table}

\begin{table}[H]
	\centering
	\caption{Basic Settings of Signalized Intersections}
	\label{tab:table00}
	\begin{tabular}{@{}lcc@{}}
		\toprule
		Terms                       & \multicolumn{1}{l}{Through} & \multicolumn{1}{l}{Left (if  applicable)} \\ \midrule
		Unit of Green Extension (s) & 2                           & 2                                         \\
		Minimum Green (s)           & 15                          & 5                                         \\
		Maximum Green (s)           & 60                          & 25                                        \\
		Yellow (s)                  & \multicolumn{2}{c}{3}                                                   \\
		Red Clearance (s)           & \multicolumn{2}{c}{3}                                                   \\
		Cycle Length (s)            & \multicolumn{2}{c}{50 to 120}                                           \\ \bottomrule
	\end{tabular}
\end{table}

In both experiments, all vehicles are set to be homogeneous. The car-following behavior is modeled with the Intelligent Driver Model \cite{Treiber2000}, and the lane-changing behavior is modeled with the SUMO built-in dynamics models \cite{Erdmann2015}. The maximum speed of all vehicles is 50 mph, and the maximal acceleration and deceleration rates are 3.3 feet/$s^2$ and 15 feet/$s^2$, respectively. In the Grid experiment, traffic flows are generated according to Poisson distribution from the intermediate nodes at the edge of each direction and merge into the four corner nodes. The volume of each traffic flow is constant and equal to 750 pcu/h. The OD matrices are first randomly generated, and then stored and used without any other further changes in the subsequent experiment. In the MoCo experiment, in order to test the robustness and effectiveness of the proposed KS-DDPG algorithm, the traffic demand data consisting of both congestion generation and recovery processes are used as the input to simulate the recurrent congestion situation. The volume increased from almost 332 pcu/h/ln to 694 pcu/h/ln and then gradually recovered to 358 pcu/h/ln at the end of simulation time. The daily Origin-Destination (OD) matrices are extracted from the Montgomery-National Capital Park and Planning Commission, and the traffic volumes are obtained from the Maryland Department of Transportation and utilized to estimate the OD matrix from the raw daily OD matrices. For the unsignalized intersections, the default stop-sign control logic in SUMO is utilized. In addition, other basic settings of all signalized intersections in both experiments are listed in Table  \ref{tab:table00}.
The KS-DDPG algorithm is trained in episodes, and different random seeds are used for generating different training and evaluation episode, but the same seed is shared for the same episode. The KS-DDPG consists of two networks, i.e., the actor network and the critic network. The actor network receives the state and the obtained knowledge and outputs the actions. A one-layer network with  $2^{9}$  units is used for observation embedding, and a one-layer network with $2^{8}$ units is used for actor selector. We used networks with three hidden layers ($2^{10}$, $2^{9}$ and $2^{8}$ units) for critics, and networks with two hidden layers ($2^{9}$ and $2^{8}$ units) for the actors. The Adam optimizer with a learning rate of 0.001 for critic and 0.0001 for policies are selected. The reward discount factor is set to be 0.95, the size of replay buffer is $10^{6}$, and the batch size is $2^{10}$. We update network parameters after every 100 samples added to the replay buffer using soft updates with $\tau=0.01$.

\subsection{Benchmarks}
The performance of the proposed algorithm is compared with the fixed-time traffic control, a state-of-the-art network-wide signal control method named MaxPressure \cite{Varaiya2013}, and three well-trained benchmark RL algorithms in a simulation replication with the same random state, including DQN introduced in Section \ref{qlearning}, DDPG introduced in Section \ref{ddpg}, and MADDPG introduced in Section \ref{maddpg}, respectively. Webster's method is used to determine the fixed-time control plan for each intersection \cite{Webster1958}. For the MaxPressure approach, we follow its original setting, which is to greedily choose the phase that maximizes the pressure (i.e., a metric related to the upstream and downstream queue lengths) \cite{Varaiya2013}. In both DQN and DDPG, the agents are trained independently. In other words, each intersection is controlled by an agent. The agent does not share observations and parameters with others, but independently updates their own networks. For the MADDPG algorithm, all settings except for knowledge sharing related parameters are the same as KS-DDPG. In addition, different hyperparameters for benchmarks are tested several times to ensure adequate performance, and fine-tuned values are injected into these models. Finally, all algorithms are trained with 1,200 episodes given episode horizon $T=720$ steps. All the computations are performed in an NVIDIA DGX-2 system.

\subsection{Experiment Results and Discussions}
\subsubsection{Convergence Comparison}
The average reward per agent per training episode of each algorithm which quantifies how well the task has been solved is calculated as

\begin{equation}\label{Eq20}
	\overline{R}=\dfrac{1}{T}\sum_{t=0}^{T-1}(\dfrac{1}{N}\sum_{\forall i\in N}r_{i,t})
\end{equation}

\begin{figure}[H]
	\centering
	\subfigure[Grid Experiment]{
		\includegraphics[width=1\linewidth]{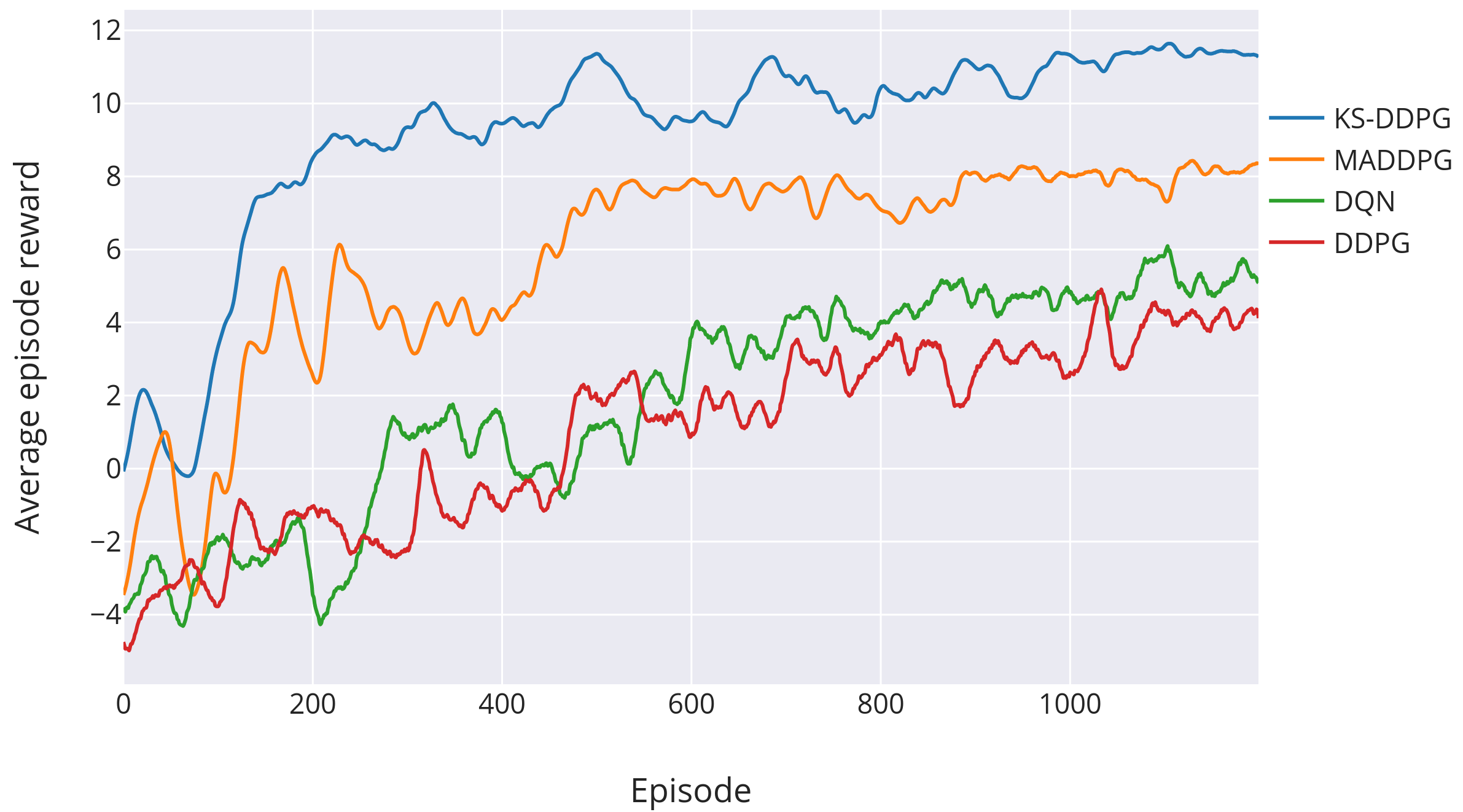}
		\label{fig:reward01}
	}
	\quad
	\subfigure[MoCo Experiment]{
		\includegraphics[width=1\linewidth]{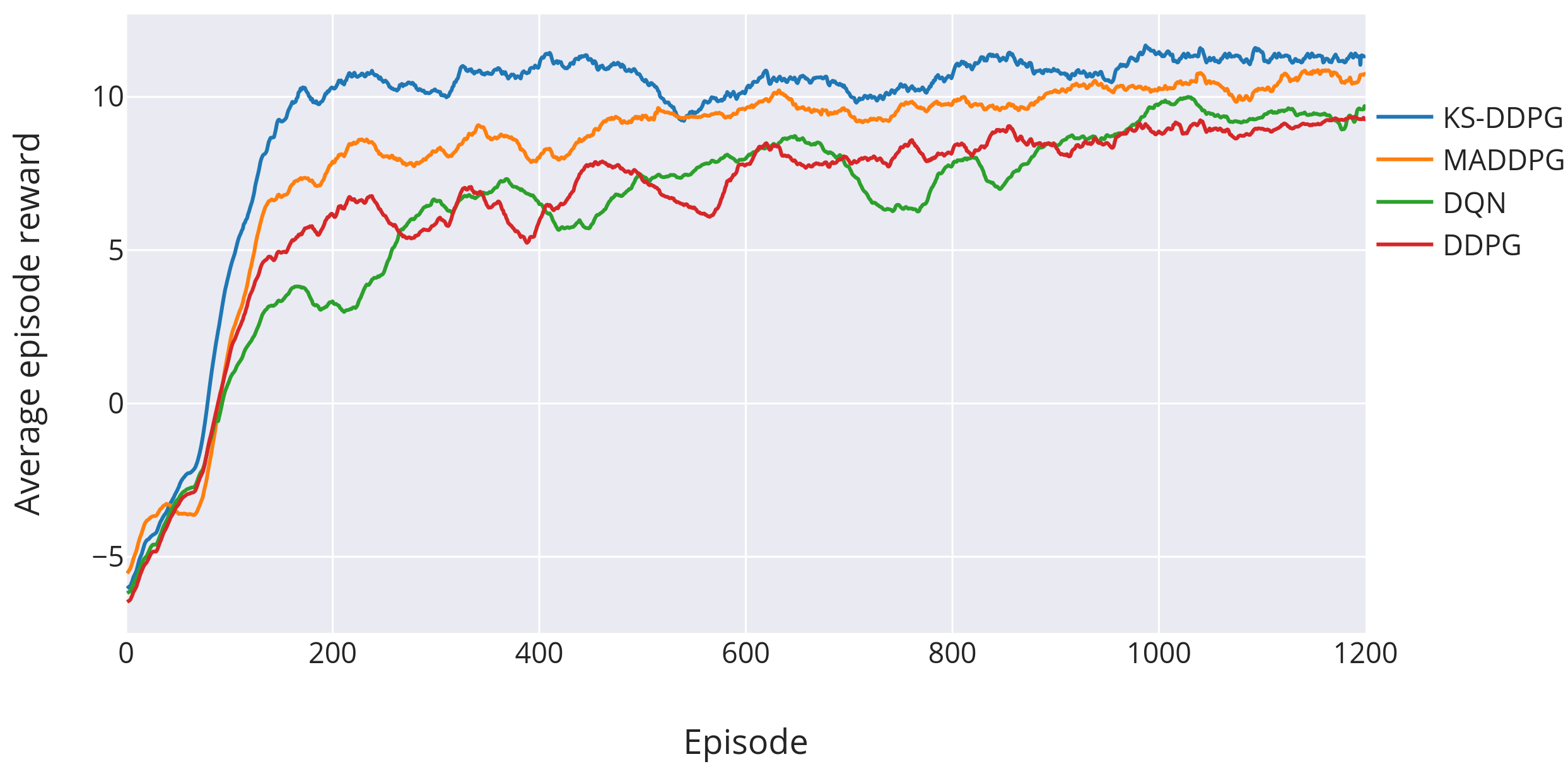}
		\label{fig:reward02}
	}

	\caption{Training Curves for KS-DDPG and Other RL Baselines}
	\label{fig:reward}
\end{figure}

The convergence comparison results calculated by Eq. \ref{Eq20} between KS-DDPG and other RL-based baselines for both experiments are demonstrated in Fig. \ref{fig:reward}. In general, as RLs learn from accumulated experience and eventually achieves a local optimum, the training curve will increase and then converge. It should be noted that the presented curves are the best ones selected from all tests, and they are smoothed with a moving average of 5 points.

From Fig. \ref{fig:reward01}, we can find that for the large-scale Grid experiment, the two independently trained RLs, i.e., DQN and DDPG, fail to learn very stable policies in given 1,200 episodes.  Although the curves have shown some promising trends,  given the great computational costs of running the full set of simulations, it was not affordable to let it run indefinitely, therefore, we only presented the convergence results within 1,200 episodes. In contrast,  the two multi-agent algorithms, MADDPG and KS-DDPG, both successfully converge as their rewards fluctuate in very small ranges at the end of the training. Moreover, compared to MADDPG, KS-DDPG presents a more superior learning curve.

Nevertheless, for the small-scale MoCo experimental results presented in Fig. \ref{fig:reward02},  the difference between learning curves of KS-DDPG and MADDPG is not as significant as in the Grid experiment. It may indicate that the policy inferring mechanism of MADDPG could yield acceptable performance when the number of agents is not too large. In addition, although not very stable, DQN and DDPG are also able to obtain positive rewards, implying that they have learned favorable policies to mitigate delays during the training process.

As the MADDPG algorithm is what our method builds on, the comparison between KS-DDPG and MADDPG (in which agents cannot communicate with others) has great potential to better demonstrate the improvements brought by the proposed commutation protocol.  Based on several benchmark empirical performance metrics, including \textit{jumpstart performance},  \textit{asymptotic performance}, and \textit{time to threshold} $\footnote{\textit{Jumpstart performance}: The initial performance after the first episode; \textit{Asymptotic performance}: The final learned performance after total episodes; \textit{Time to threshold}: The learning time needed to achieve a prespecified performance level.}$  \cite{Taylor2009}, we can discover that KS-DDPG performs betters than MADDPG in both experiments by a huge amount. These results indicate that the knowledge-sharing mechanism does not retard model convergence, but on the contrary speeds up the search for optimal policy. Besides, these results also imply that this explicit communication technique could help MARL learn better policies more effortlessly in complex environments.

\subsubsection{Control Performance Comparison}

\begin{figure}[H]
	\centering
	\subfigure[Grid Experiment]{
		\includegraphics[width=1\linewidth]{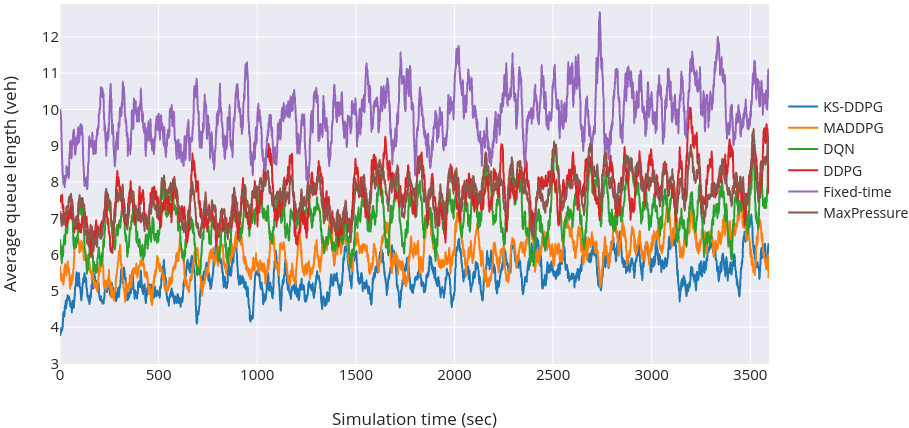}
		\label{fig:queue01}
	}
	\quad
	\subfigure[MoCo Experiment]{
		\includegraphics[width=1\linewidth]{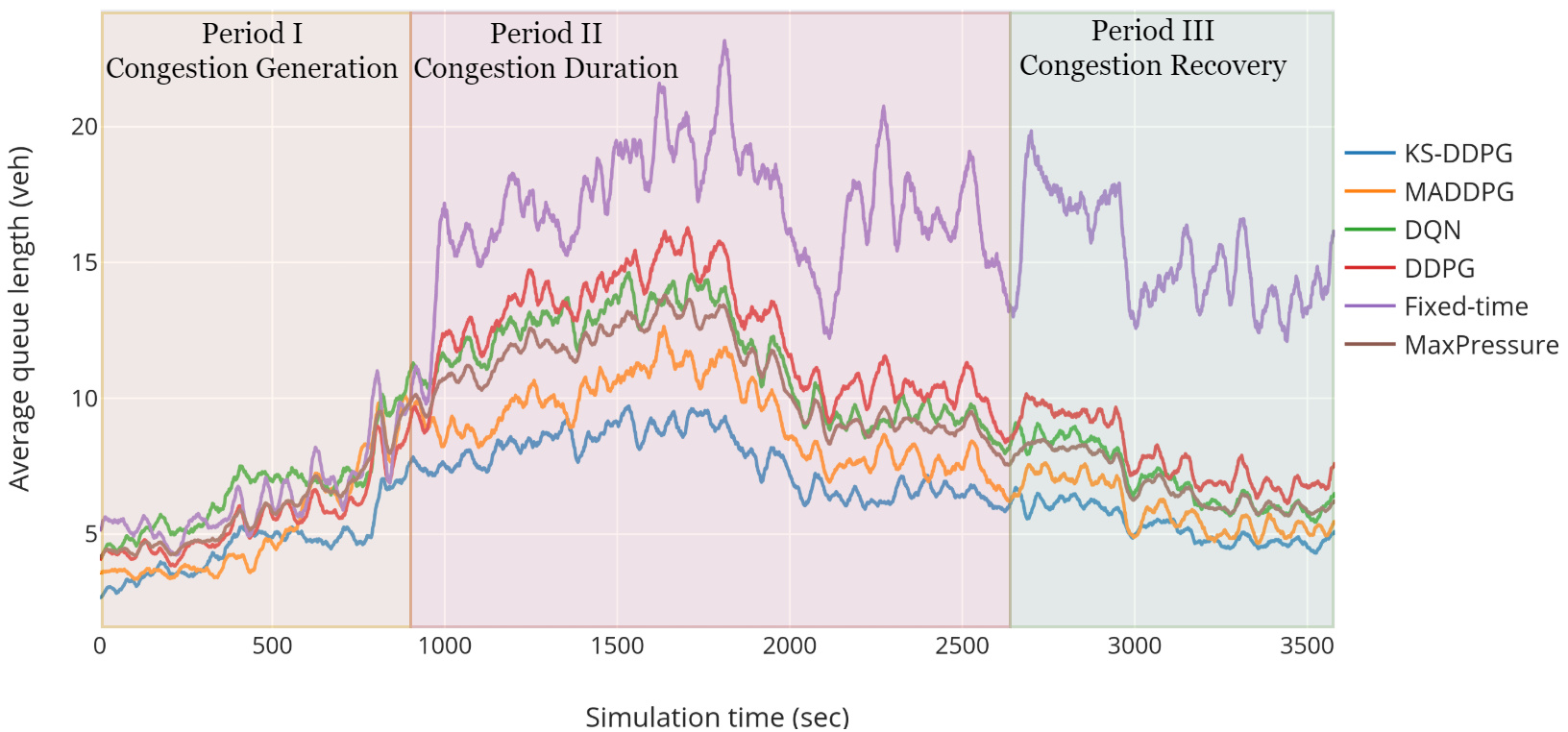}
		\label{fig:queue02}
	}
	\caption{Average Queue Length for KS-DDPG and Other Baselines}
	\label{fig:queue}
\end{figure}

\begin{table}[H]
	\centering
	\caption{Control Performance Comparison in Grid Experiment}
	\label{tab:table01}
	\resizebox{\textwidth}{!}{%
		\begin{tabular}{@{}lllllll@{}}
			\toprule
			\textbf{Metrics} & \textbf{KS-DDPG} & \textbf{MADDPG} & \textbf{DQN} & \textbf{DDPG} & \textbf{Fixed-time} & \textbf{MaxPressure} \\ \midrule
			Avg. Reward                     & 9.20 & 5.84  & 1.88  & 1.05  & NA    & NA    \\
			Avg. Queue Length {[}veh{]}     & 5.45  & 6.03  & 7.22  & 7.67  & 9.82  & 7.66  \\
			Avg. Intersection Delay {[}s{]} & 30.74 & 35.73 & 45.71 & 47.10 & 62.21 & 47.35 \\
			Avg. Vehicle Speed {[}feet/s{]} & 22.34 & 20.59 & 18.25 & 19.03 & 13.77 & 18.88 \\
			Avg. Number of Stops {[}/s{]}   & 6.28  & 6.73  & 7.99  & 8.02  & 11.35 & 7.90  \\ \bottomrule
		\end{tabular}%
	}
\end{table}

\begin{table}[H]
	\centering
	\caption{Control Performance Comparison in MoCo Experiment}
	\label{tab:table02}
	\resizebox{\textwidth}{!}{%
		\begin{tabular}{@{}lllllll@{}}
			\toprule
			\textbf{Metrics} & \textbf{KS-DDPG} & \textbf{MADDPG} & \textbf{DQN} & \textbf{DDPG} & \textbf{Fixed-time} & \textbf{MaxPressure} \\ \midrule
			Avg. Reward                               & 9.47  & 8.17  & 6.31  & 6.73  & NA     & NA    \\
			Avg. Queue Length in Period I {[}veh{]}   & 4.57  & 5.26  & 6.60  & 5.54  & 6.35   & 5.93  \\
			Avg. Queue Length in Period II {[}veh{]}  & 6.80  & 8.71 & 9.65 & 9.01 & 12.62  & 8.88 \\
			Avg. Queue Length in Period III {[}veh{]} & 5.47  & 6.28  & 7.26 & 8.20  & 13.55  &7.13  \\
			Avg. Queue Length  {[}veh{]} & 6.27  & 7.34  &9.03 & 9.44  & 13.85 &8.53  \\
			Avg. Intersection Delay {[}s{]}           & 41.75 & 50.33 & 62.72 & 65.14 & 93.27 & 62.04 \\
			Avg. Vehicle Speed {[}feet/s{]}           & 20.45 & 18.02 & 16.20 & 16.16 & 13.55  & 17.23 \\
			Avg. Number of Stops {[}/s{]}             & 11.76 & 13.82 & 15.47 & 16.06 & 24.35  & 16.22 \\ \bottomrule
		\end{tabular}%
	}
\end{table}

The performance comparison between KS-DDPG and other baselines with reference to the average queue length of the entire network at each simulation step in the Grid experiment is presented in Fig. \ref{fig:queue01}. It is noteworthy that the obvious variability of these curves is caused by phase switching. It can be observed that all the RL-based methods have smaller average queue lengths than the fixed-time control, and KS-DDPG outperforms others as expected. In addition, both DQN and DDPG have very similar performance with the state-of-the-art model-driven traffic control optimization approach, i.e., MaxPressure. These results demonstrate that all RL-based control methods, especially KS-DDPG, have certain improvements in reducing intersection congestion compared with traditional methods. Table \ref{tab:table01} further demonstrates the comprehensive comparison results these control methods. These metrics in Table \ref{tab:table01} are firstly spatially aggregated at each time over evaluation episodes, then the temporally averaged results are calculated and presented. As the results show, KS-DDPG achieves significant performance improvements over all the baselines. Compared with MaxPressure, on average,  KS-DDPG reduces about 28.9\% queue length, 35.1\% intersection delay,  and 21.0\% number of stops, and increases about 18.3\% vehicle speed. KS-DDPG also has about at least 10\% improvement on each measurement than the MADDPG control method.

Similarly, the evaluation results on queue lengths of the MoCo experiment are presented in Fig. \ref{fig:queue02}. Besides, the detailed comparisons on control metrics of these methods are listed in Table \ref{tab:table02}. Different from that in the Grid experiment, based on the V/C ratio, we roughly divide the whole simulation process into three periods, including congestion generation,  congestion duration, and congestion recovery. It can be discovered that when the traffic volume is not large (V/C $< $0.5), all the methods have acceptable control performance. When the volume increases sharply in Period II, the average queue lengths of different methods also grow significantly, and pretty noticeable differences of them could be perceived in the figure. During this period, the maximum average queue length of the fixed time control has reached above 25 veh/ln, indicating the network is highly congested. In contrast, the KS-DDPG has a more favorable control outcome, as its maximum average queue length is less than 10 veh/ln. The control performance of other RLs are all better than the fixed time but worse than KS-DDPG in Period II, and MADDPG performs the best among these three RLs. Besides, MaxPressure performs better than DQN and DDPG,  but the difference is not very significant. In Period III, all RLs and MaxPressure demonstrate a significant trend to alleviate congestion. However, for fixed-time, this trend is not apparent. As shown in the green area of  Fig. \ref{fig:queue02}, KS-DDPG can promptly alleviate congestion as expected, while the average queue lengths of other RLs rebounded to some extent with the fluctuation of traffic. Interestingly, as shown in Table \ref{tab:table02}, the performance gaps between KS-DDPG and other baselines become significantly larger than those in the Grid experiment. For instance, in the Grid experiment, the number of stops for MADDPG, DQN, DDPG, fixed-time, and MaxPressure are 1.07, 1.27, 1.28, 1.81, and 1.26 times than that for KS-DDPG, respectively. However, these numbers in the MoCo experiment became 1.17, 1.32, 1.37, 2.01, and 1.38, respectively. Similar findings can also be observed in the comparison of these methods in other metrics.  All in all, the above results prove that KS-DDPG has a sound control outcome, no matter during the periods of congestion generation, duration or recovery.

In conclusion, compared with conventional transportation and RL-based baselines, KS-DDPG achieves consistent performance improvements across different road networks and traffic conditions. The advantage becomes even larger as the assessed traffic scenario changes from synthetic regular network to real-world irregular and dynamic environment. We believe these favorable results are attributed to the explicit communication among agents enabled by the knowledge-sharing mechanism. More specifically, since fixed-time and MaxPressure controls are not able to learn from the feedback of the environment, it is not surprising that they can barely achieve satisfactory results. DQN and DDPG also show limited capabilities in alleviating congestion, especially in dynamic traffic situations, mainly because they independently optimize the strategy of a single intersection. Furthermore, our approach also has obtained at least 10\% control improvements over the state-of-the-art multi-agent RL, i.e., MADDPG. Since there are no communication channels in MADDPG, agents are unable to share their understanding of the environment during modeling. Instead, this algorithm utilizes a probabilistic network and maximizes the log probability of outputting another agent's observed actions to infer other agents' policies. The growing performance gaps between KS-DDPG and MADDPG from the Grid experiment to the MoCo experiment also demonstrate that direct cooperation through knowledge sharing is better than implicit cooperation through inference of other policies.

\begin{table}[H]
	\centering
	\caption{Training time for different models}
	\label{tab:my-table0}
	\begin{tabular}{@{}lllll@{}}
		\toprule
		\multirow{2}{*}{Experiment} & \multicolumn{4}{l}{Avg. training time for one episode (minutes)} \\ \cmidrule(l){2-5}
		& KS-DDPG & MADDPG & DQN    & DDPG   \\ \midrule
		Grid & 36.37   & 30.86  & 125.28 & 111.11 \\
		MoCo & 5.22    & 5.13   & 8.13   & 7.85   \\ \bottomrule
	\end{tabular}
\end{table}

In addition, we also compare the training time of KS-DDPG and other RL-based baselines, and the results are presented in Table \ref{tab:my-table0}. In order to keep fair comparisons, all the methods are individually evaluated on the computing platform.  It can be discovered that the training time for the proposed KS-DDPG is comparable to that of MADDPG in both experiments, indicating the extra computational burden brought by the communication channel is acceptable.

\section{Conclusions}
In this work, we have introduced KS-DDPG, a novel multi-agent reinforcement learning framework, to optimize network-wide ATSC. A knowledge-sharing mechanism is utilized as the intra-agent communication protocol to improve coordination skills among agents. The knowledge stored in the knowledge container is a representation of the collective learning of the traffic environment collected by all agents, and is used to better form individual policies for each agent. The proposed algorithm is compared with both state-of-the-art conventional transportation and RL-based methods by careful evaluation in synthetic and real-world scenarios. Results show that KS-DDPG achieves superior performance relative to all the baselines with acceptable computational costs.  Compared with MADDPG, the significantly more stable training curves and favorable traffic control results strongly prove the importance of introducing explicit communication channels to achieve agent cooperation.

One of the main limitations of the algorithm is that all agents need to perform communication during the entire modeling process, resulting in limited overall communication efficiency. One possible solution is to add a “master agent” to determine the appropriate time for each agent to access the knowledge container. Although this measure adds an extra layer of complexity to the model, it may reduce the number of knowledge container accesses required in a large-scale system and increase overall scalability. Besides, in future work, we will consider using heterogeneous vehicles to build a more realistic traffic flow. We will also try to optimize traffic control and route guidance coordinately under the mixed flow consisting of both connected and conventional vehicles.

\section*{Acknowledgments}
This paper is supported by National Key Research and Development Program of China (No. 2019YFB2102100), The Science and Technology Development Fund of Macau SAR (File no. 0015/2019/AKP),  and Guangdong-Hong Kong-Macao Joint Laboratory of Human-Machine Intelligence-Synergy Systems (No. 2019B121205007).

\end{document}